\documentclass{article}
\usepackage[margin=1.25in]{geometry}

\usepackage{amsfonts}       
\usepackage{nicefrac}       

\usepackage{algorithm}
\usepackage[noend]{algpseudocode}
\usepackage{enumitem}
\usepackage{caption}
\usepackage{subcaption}

\usepackage{amsmath}
\usepackage{amssymb}
\usepackage{mathtools}
\usepackage{amsthm}

\theoremstyle{plain}
\newtheorem{theorem}{Theorem}[section]

\newtheorem{lemma}[theorem]{Lemma}
\newtheorem{corollary}[theorem]{Corollary}
\theoremstyle{definition}

\usepackage{makecell}
\newcommand{\bp}{\mathbf{p}}
\newcommand{\bX}{\mathbf{X}}
\newcommand{\eps}{\epsilon}
\newcommand{\heta}{\hat{\eta}}
\newcommand{\goes}{\rightarrow}
\newcommand{\parent}{\mathrm{pa}}
\newcommand{\hv}{\hat{v}}
\newcommand{\hH}{\hat{H}}
\newcommand{\bac}{\bar{c}}
\newcommand{\baG}{\bar{G}}
\newcommand{\hC}{\hat{C}}
\newcommand{\hG}{\hat{G}}
\newcommand{\hI}{\hat{I}}
\newcommand{\tG}{\tilde{G}}
\newcommand{\CC}{\mathcal{C}}
\newcommand{\Algo}[1]{\textsc{#1}}
\newcommand{\given}{\, | \,}
\renewcommand{\vec}[1]{\boldsymbol{#1}}
\newcommand{\bx}{\vec{x}}
\DeclareMathOperator*{\argmax}{\arg \max}

\providecommand{\inner}[1]{\left\langle#1\right\rangle}

\DeclareMathOperator{\E}{E}
\allowdisplaybreaks

\title{Private and Communication-Efficient Algorithms for Entropy Estimation\footnotetext{Originally published at the 36th Conference on Neural Information Processing Systems (NeurIPS 2022). This version corrects some errors in the original version.}}

\author{%
  Gecia \mbox{Bravo-Hermsdorff} \\
  Department of Statistics\\
  University College London \\
  \texttt{gecia.bravo@gmail.com} \\
  \and
   R\'obert \mbox{Busa-Fekete}\\
   Google Research \\
   \texttt{busarobi@google.com} \\
  \and
   Mohammad Ghavamzadeh \\
   Google Research \\
   \texttt{ghavamza@google.com} \\
  \and
   Andres Mu\~noz Medina\\
   Google Research \\
   \texttt{ammedina@google.com} \\
  \and
   Umar Syed \\
   Google Research \\
   \texttt{usyed@google.com} \\
}
\date{}

\begin{document}

\maketitle

\begin{abstract}
Modern statistical estimation is often performed in a distributed setting where each sample belongs to a single user who shares their data with a central server. Users are typically concerned with preserving the privacy of their samples, and also with minimizing the amount of data they must transmit to the server. We give improved private and communication-efficient algorithms for estimating several popular measures of the entropy of a distribution. All of our algorithms have constant communication cost and satisfy local differential privacy. For a joint distribution over many variables whose conditional independence is given by a tree, we describe algorithms for estimating Shannon entropy that require a number of samples that is linear in the number of variables, compared to the quadratic sample complexity of prior work. We also describe an algorithm for estimating Gini entropy whose sample complexity has no dependence on the support size of the distribution and can be implemented using a single round of concurrent communication between the users and the server. In contrast, the previously best-known algorithm has high communication cost and requires the server to facilitate interaction between the users. Finally, we describe an algorithm for estimating collision entropy that generalizes the best known algorithm to the private and communication-efficient setting.
\end{abstract}

\section{Introduction}

Statistical estimation has traditionally focused on minimizing the number of samples needed to estimate properties of a distribution. In the `big data' era, statisticians and computer scientists have also tried to minimize the space complexity of estimation algorithms, particularly in the streaming setting. More recently, the increasing prevalence of mobile computing has led to a focus on the privacy and communication costs of statistical estimation. In this paper, we consider the following setting: a set of users each draw one sample from a distribution, and share information about their sample with a central server. The central server then uses the collected data to estimate a property of the distribution. Users are concerned with preserving the privacy of their sample, and also with minimizing the amount of data that is transmitted to the server.

For example, consider the problem of detecting fingerprinting on the web. Many websites track users across the web without their consent by recording (enough) information about their devices (e.g., installed fonts, operating system, timezone, etc.), a practice known as ``browser fingerprinting''. Entropy is the standard metric used to quantify the identifiability of the collected fingerprints. So a private and distributed method for estimating entropy can be used by a browser to warn users that this covert tracking could occur, without ever storing the fingerprints themselves.

The study of entropies has an extensive and rich history in mathematics and sciences. Related quantities called “entropy” appear in many contexts (thermodynamics, information theory, dynamical systems \cite{n7}, category theory \cite{n6}, etc.). These may be broadly thought as measures of information of a system or process obeying certain properties, which, in turn, lead to natural measures of disorder, randomness, outcome diversity, information content, uniformity, etc.

In this paper, we study private and communication-efficient algorithms for estimating certain entropies of a distribution. Specifically, we give algorithms for estimating the following entropies, which are widely-used in many scientific fields to quantify the uncertainty, diversity and spread of a discrete distribution:
\begin{itemize}
    \item \textbf{\textit{Shannon entropy}} \cite{n5}, a fundamental quantity in information theory.
    \item \textbf{\textit{Gini entropy}} (also known as  Tsallis entropy \cite{n3} of order 2,  or (one minus the) second frequency moment).
    Some of its applications include measuring
    ecological diversity \cite{simpson1949measurement,n1}, market competition among firms \cite{Herfindahl1950ConcentrationIT}, effective size of political parties \cite{LaaksoTaagepera}, and suitability of features to split on during decision tree learning~\cite{raileanu2004theoretical}.
    \item \textbf{\textit{Collision entropy}} (also known as Rényi entropy \cite{n4} of order 2).
    Some of its applications include measuring the quality of random number generators \cite{TRNG}, and determining the number of reads needed to reconstruct a DNA sequence \cite{motahari2013information}.
\end{itemize}

Our algorithms are implemented in either the \textit{non-interactive} model (for the Gini and collision entropies), in which all users simultaneously
exchange information with the server during a single round of communication, or the (stronger)
\textit{sequentially interactive} model (for the Shannon entropy), in which the server queries users one at a time, possibly in an adaptive
manner~\cite{joseph2019role}. 
When analyzing the communication complexity of an algorithm, we prove bounds on the number of bits that each user transmits to the server. 
However, the server is allowed to broadcast an arbitrary amount of information to the users (this is also called the \textit{blackboard} model \cite{duchi2019lower}), including shared random bits (also known as the \textit{public coin} model \cite{pmlr-v89-acharya19b,pmlr-v99-acharya19a,JosephMN019}). 
When analyzing the privacy of our algorithms we use the framework of \textit{local differential privacy} \cite{dwork2014algorithmic}, which ensures that the server learns very little about each user’s data.

\subsection{Our contributions}

\begin{itemize}[leftmargin=*]
    \item A sequentially interactive $\alpha$-local differentially private algorithm for estimating the Shannon entropy of a joint distribution on $d$ variables within $\eps d$ error using $\tilde{O}(d / (\alpha^2 \epsilon^5))$ samples and $O(1)$ bits per sample. Our analysis assumes that each of the $d$ variables has a constant support size and that their conditional independence graph is a tree. We also describe algorithms that have better dependence on $1/\eps$ in certain special cases, such as when the tree has low diameter or is a chain. Our algorithms achieve $O(1)$ communication complexity by observing only two or three of the $d$ variables in any single sample; we call these \emph{pair} and \emph{triplet observations}. The only previously known algorithm for estimating the Shannon entropy of a tree-structured distribution from \emph{pair} observations is a non-interactive algorithm due to Chow and Liu \cite{ChLi68}. We prove that any non-interactive algorithm requires $\Omega(d^2)$ \emph{pair} observations to achieve $O(d)$ error. We also prove that, for any sequentially interactive algorithm, $\Omega(d / \eps)$ pair observations are necessary to achieve $O(\eps d)$ error.
    \item A non-interactive $\alpha$-local differentially private algorithm for estimating the Gini entropy of a distribution within $\eps$ error using $O(1 / (\alpha^4 \eps^2))$ samples, $O(1)$ bits per sample, and $\tilde{O}(1)$ space. The best previous algorithms \cite{butucea2021locally} either have a sample complexity that depends on the support size $k$ of the distribution, or are sequentially interactive, and also require $\Omega(k)$ bits per sample and $\Omega(k)$ space. Also our error bound holds with high probability instead of only in expectation.
    \item A non-interactive $\alpha$-local differentially private algorithm for estimating the collision entropy of a distribution with support size $k$ within $\eps$ error using $\tilde{O}(k^2 / (\alpha^4\eps^2 ))$ samples, $O(1)$ bits per sample, and $\tilde{O}(1)$ space. Our algorithm generalizes the previously best known non-interactive algorithm  \cite{TRNG} to the private and communication-efficient setting.
\end{itemize}

\section{Related Work}

There is a very extensive literature on distributed statistical estimation under communication constraints (see \cite{zhang2013information} for the paper that appears to have started this thread). Variations on the problem include whether communication is allowed between users, whether communication happens in one or multiple rounds, whether there is a shared source of randomness among the users, and whether communication is limited per-user or only cumulatively across all users.

Many previous results in this area bound the sample and communication complexity of estimating the parameters of a distribution $P_\theta$, where $\theta \in \Theta$ (see e.g.~\cite{han2018geometric}). This problem class includes discrete distribution estimation, where the guarantees are usually stated as bounds on the relative entropy or total variation distance between the estimated and true distribution (see e.g.~\cite{acharya2019hadamard}). Other problems of interest are mean estimation \cite{suresh2017distributed} and heavy hitter estimation \cite{acharya2019communication}.

There has also been significant interest in differentially private statistical estimation, and of particular relevance is the work by \cite{acharya2018inspectre}, who gave private algorithms for estimating certain functionals of a distribution, including the Shannon entropy. However, they used the central model of differential privacy, while in this paper we prove guarantees using the (stronger) local model.

\section{Entropy Measures}
\label{sec:entropy}

The \emph{Shannon}, \emph{Tsallis}, and \emph{R\'enyi} entropy of a discrete random variable $X$ are defined as
\begin{align}
\label{eq:Shannon-def}
\hspace{-0.75in}\text{(Shannon)} \qquad\qquad H(X) &= -\sum_x \Pr[X = x] \log \Pr[X = x], \\
\label{eq:Tsallis-def}
\hspace{-0.75in}\text{(Tsallis)} \qquad\qquad T_\gamma(X) &= \frac{1}{\gamma - 1} \big( 1 - \sum_x \Pr[X = x]^\gamma \big), \\
\label{eq:Renyi-def}
\hspace{-0.75in}\text{(R\'enyi)} \qquad\qquad R_\gamma(X) &= \frac{1}{1 - \gamma} \log\big(\sum_x (\Pr[X = x])^\gamma\big),
\end{align}
where $\gamma$ in~\eqref{eq:Tsallis-def} and~\eqref{eq:Renyi-def} is a free parameter satisfying $\gamma > 0$ and $\gamma \neq 1$. Both Tsallis and R\'enyi entropy are generalizations of Shannon entropy in the sense that $\lim_{\gamma \goes 1} T_\gamma(X) = \lim_{\gamma \goes 1} R_\gamma(X) = H(X)$.

In this paper, we describe algorithms for estimating the Shannon entropy and special cases of the Tsallis and the R\'enyi entropy that are widely used in many scientific fields: $T_2(X)$, also known as the \emph{Gini entropy}, and $R_2(X)$, also known as the \emph{collision entropy}. Substituting $\gamma = 2$ into the definitions above and using the abbreviations $G(X) \equiv T_2(X)$ and $C(X) \equiv R_2(X)$, we have:
\begin{align*}
\hspace{-0.75in}\text{(Gini)} \qquad\qquad G(X) &\equiv T_2(X) = 1 - \sum_x \Pr[X = x]^2,\\
\hspace{-0.75in}\text{(Collision)} \qquad\qquad C(X) &\equiv R_2(X) = -\log\big(\sum_x \Pr[X = x]^2\big).
\end{align*}
Gini entropy is so-called because it is equivalent to the Gini diversity index, a statistic proposed by Corrado Gini in 1912 to measure income and wealth inequality \cite{n8}. Collision entropy takes its name from the observation that if $X$ and $X'$ are independent and identically distributed, then $C(X) = -\log \Pr[X = X']$.

For the problem of estimating Shannon entropy, we specialize to a high-dimensional setting, where we only observe a \textit{pair} (or \textit{triplet}) of the dimensions at a time. That is, \(X\) is a random-vector of \(d\) discrete variables, where \(d\) is large, but each \(X_i\) has a constant support size (e.g., they are binary), and we only observe two (or three) dimensions per sample.  Without making any assumption about this joint distribution, the problem is intractable. One of the most common assumptions, which we also adopt in this work, is that the joint distribution is tree-structured. In this case, the distribution can be estimated by the celebrated \cite{ChLi68} (and optimal \cite{bhattacharyya2021near}) Chow-Liu algorithm. 
While the Chow-Liu algorithm requires \(\Omega(d^2)\) \textit{pairs} observations to estimate the Shannon entropy, our sequential algorithm requires only \(\mathcal{O}(d)\) \textit{pairs} observations (see Section~\ref{sec:gentree} for more details).

The \emph{joint Shannon entropy} $H(X_1, \ldots, X_d)$ of a set of random variables $X_1, \ldots X_d$ is the Shannon entropy $H(X)$ of the random variable \mbox{$X = (X_1, \ldots, X_d)$}. 
We write the abbreviated term \emph{joint entropy} when the use of Shannon entropy is obvious from context.

The \emph{mutual information} between two random variables $X$ and $Y$ and their \emph{conditional mutual information} given another random variable $Z$ are defined as: 
\begin{align}
I(X; Y) &= H(X) + H(Y) - H(X, Y), \label{eq:mut1} \\
I(X; Y ~|~ Z) &= H(X, Z) + H(Y,Z) - H(X,Y,Z) - H(Z). \label{eq:mut2}
\end{align}

\section{Estimation Algorithms and Evaluation Criteria}

A set of $n$ users and a central server cooperate according to the following protocol to estimate the entropy of a random variable $X$:

\begin{enumerate}[leftmargin=*]
    \item Each user $i \in [n]$ draws an independent sample $x_i$ according to the distribution of $X$.
    \item For $r$ rounds:
    \begin{enumerate}
        \item The server sends information to a subset of the users.
        \item Those users send (partial) information about their sample back to the server.
    \end{enumerate}
    \item The server outputs an estimate of the Shannon entropy (Algorithms~\ref{alg:fastmst},~\ref{alg:markov_chain}, and~\ref{alg:star}) or the Gini or collision (Algorithm~\ref{alg:ginicollision}) entropies of $X$.
\end{enumerate}

An \emph{estimation algorithm} specifies the steps that each user and the server perform to implement the above protocol. The algorithm is \emph{non-interactive} if the protocol consists of a single round in which all users participate. In a non-interactive algorithm the server cannot adapt its queries to users based on responses from other users, since the server communicates with all the users concurrently. An algorithm is \emph{sequentially interactive} if each round consists of communication with a single user, who is never contacted again. Sequential interactivity enables the server to query users adaptively \cite{joseph2019role}.

We evaluate estimation algorithms according to the following criteria:
\begin{itemize}[leftmargin=*]
  \item \emph{Sample complexity}: The number of users from whom the server requests data.
  \item \emph{Space complexity}: The space used by the server when executing the algorithm.
  \item \emph{Communication complexity}: The maximum number of bits transmitted by any single user to the server. Note that the amount of information sent by the server to the users is not counted when determining communication complexity.
  \item \emph{Privacy}: Let $x_i$ be the sample belonging to user $i$ and $o_i$ be the data observed by the server from user $i$. We say that an algorithm satisfies \emph{$\alpha$-local differential privacy} if
  \[
  \Pr[o_i \in O ~|~ x_i = x] \le e^\alpha \Pr[o_i \in O ~|~ x_i = x']
  \]
  for any user $i$, measurable set $O$, and possible sample values $x, x'$.
  \item \emph{Error}: The absolute difference between the true entropy of the distribution and the estimate output by the server.
\end{itemize}

\section{Estimating Shannon Entropy of Tree-structured Joint Distributions}

In this section we assume that \mbox{$X = (X_1, \ldots, X_d)$} is a vector of $d$ discrete variables, and that the support size of each variable $X_i$ is constant (e.g., each variable is Boolean). We also assume that $X$ has a \emph{tree-structured} distribution, which means that there exists a rooted tree $T$ with $d$ nodes such that for any node $i \in [d]$ we have
$
\Pr[X_i \mid X_{-i}] = \Pr[X_i \mid X_{\parent_T(i)}],
$
where $X_{-i} = (X_1, \ldots X_{i-1}, X_{i+1}, \ldots, X_d)$ and the node $\parent_T(i)$ is the parent of node $i$ in tree $T$. If $i$ is the root node, then we define $\Pr[X_i \mid X_{\parent_T(i)}] = \Pr[X_i]$.
Equivalently, a tree-structured distribution is a Markov random field with a tree as the underlying undirected graph.
Essentially, the tree-structured assumption implies that the only correlations among the \(X_i\)’s are pairwise correlations.
If \(T\) is a chain or a star we say that \(X\) is \textit{chain-structured} and \textit{star-structured}, respectively. We will treat these two special cases at the end of this section (Algorithms~\ref{alg:markov_chain} and~\ref{alg:star}, respectively).

\subsection{Estimating Entropy of a Marginal Distribution When the Support Size is Small}
\label{sec:privateentropy}

Before proceeding to describe algorithms for estimating the Shannon entropy of tree-structured distributions, we use existing results for private distribution estimation to devise a local differentially private estimator for the Shannon entropy that is sample and communication efficient when the support size of the distribution is small (as is the case for the individual marginals). The server will repeatedly invoke this algorithm as a subroutine in the sections below.

First we recall that the difference in Shannon entropy of two random variables can be upper bounded according to Theorem 17.3.3 of \cite{Cover2006} as 
\[
| H( X_{\bp} ) - H( X_{\bp'} ) | \le \| \bp - \bp' \|_1 \log \frac{c}{\| \bp - \bp' \|_1}
\]
where $X_{\bp}$ and $X_{\bp'}$ are two discrete random variables with support size $c$ and distributions $\bp$ and $\bp'$. Next, we apply a local differentially private learning algorithm for discrete distribution due to \cite{acharya2019communication} that learns the parameters of a discrete distribution with small $L_1$ error.  The following theorem combines these two results by using the fact that \mbox{$x/\log(1/x) \le 1$} whenever \mbox{$0<x\le 1/2$}.
\begin{theorem} \label{thm:privateshannon}
For any discrete distribution $X$ with support size $c$ and for any $1/2\ge \epsilon > 0$, there exists an estimator satisfying $\alpha$-local differential privacy that estimates $H(X)$ within $\epsilon$ error using $n = O(c^2 \log\frac1\delta /(\epsilon^2 \alpha^2))$ samples with probability $1 - \delta$ when $\alpha \in (0, 1)$.
\end{theorem}

This algorithm resulting from Theorem \ref{thm:privateshannon} can be used to privately estimate the entropy $H(X_i)$,  mutual information $I(X_i; X_j)$, and  conditional mutual information $I(X_i; X_j ~|~ X_k)$ of any variables $X_i, X_j$ and $X_k$. This can be done using  \mbox{$O\left(\frac{\log\frac1\delta}{\alpha^2\eps^2}\right)$} samples per estimate and $O(1)$ bits per sample, since each of these variables has constant support size, and both mutual information and conditional mutual information can be expressed in terms of entropies (Eqs.~\eqref{eq:mut1} and \eqref{eq:mut2}). We call such an estimate \emph{$(\alpha, \eps, \delta)$-good}.

\subsection{Our Algorithm for Tree-structured Joint Distributions}
\label{sec:gentree}

Note that the support size of $X$ can be exponential in $d$. In the worst case, estimating the entropy of a distribution with support size $k$ within constant error requires $\tilde{\Theta}(k)$ samples \cite{fukuchi2017minimax}. However the tree-structure of $X$ can be exploited to significantly reduce the sample complexity. In their seminal paper, Chow and Liu~\cite{ChLi68} proved the identity
\begin{align}\label{eq:clmain}
    H(X) = \sum_{i=1}^d H(X_i) - \max_T \sum_{i=1}^d I(X_i ; X_{\parent_T(i)}),
\end{align}
for any tree-structured random variable $X$, where the maximization is taken over all possible trees connecting the $d$ variables. 

Eq.~\eqref{eq:clmain} suggests a communication-efficient algorithm for estimating the entropy of $X$, which is known as the \emph{Chow-Liu algorithm}:  First, estimate each marginal entropy $H(X_i)$ and each mutual information $I(X_i; X_j)$. Next, compute a maximum spanning tree on the $d$ variables, where the weight of each edge $(X_i, X_j)$ is the estimate of the mutual information $I(X_i; X_j)$. Finally, plug these estimators into Eq.~\eqref{eq:clmain}.

The Chow-Liu algorithm requires $\Omega(d^2)$ samples, since it computes the mutual information between every pair of variables in order to compute a maximum spanning tree. However, estimating the right-hand side of Eq.~\eqref{eq:clmain} only requires estimating the \emph{weight} of the maximum spanning tree, which is significantly easier than finding the tree itself. Algorithm \ref{alg:fastmst} adapts a technique from \cite{chazelle2001approximating} that estimates the weight of the maximum spanning tree of a graph in time that is sublinear in the number of edges in the graph. The basic idea is to select nodes of the graph at random and use breadth-first search to determine the size of each of their connected components if we were to drop edges that do not meet a weight threshold, short-circuiting the search when the size becomes too large. These quantities are combined to estimate the weight of the maximum spanning tree. In our case, an edge weight is a mutual information between a pair of variables, which we estimate from pair observations.

\begin{algorithm}
\caption{Shannon entropy estimation for tree-structured distribution\label{alg:fastmst}}
\begin{algorithmic}[1]
\State Let $M = \left\lceil \frac{2 \log c}{\eps} \right\rceil$ and $R = \left\lceil \frac{\log \frac1\delta}{\eps^2} \right\rceil$, where $c$ is an upper bound on the support size of each $X_i$.
\For{$m = 1, \ldots, M$}
\For{$r = 1, \ldots, R$}
\State Choose positive integer $Z$ randomly according to $\Pr[Z \ge z] = 1 / z$.
\State Choose $i^*$ uniformly at random from $[d]$.
\State Initialize queue to contain $i^*$ and a set $V=\{i^*\}$. 
\While{queue length is non-zero and shorter than $\min\left\{\frac2\eps, Z\right\}$} \Comment{Breadth-first search}
\State Remove $i$ from front of queue and $V = V\cup \{i\}$.
\For{$j = [d] \setminus V$}
\State Server computes $\left(\alpha, \frac{\eps}{2}, \frac{\delta}{d^2}\right)$-good estimate $\hI_{ij}$ of $I(X_i; X_j)$.
\State (Only compute this estimate once per pair.)
\If{$\hI_{ij} \ge \eps m$} add $j$ to back of queue.
\EndIf
\EndFor
\EndWhile
\If{queue length is zero} $\gamma_{mr} \gets 1$ {\bf else} $\gamma_{mr} \gets 0$.
\EndIf
\EndFor
\State $\;\heta_m \gets \frac{d}{R} \sum_{r=1}^R \gamma_{mr}$.
\EndFor
\State $\hat{W} \gets \epsilon Md - \epsilon\sum_{m=1}^M \heta_m$
\State Server computes $\left(\alpha, \eps, \frac{\delta}{d}\right)$-good estimate of each marginal entropy $H(X_i)$.
\State Let $\hat{S}$ be the sum of the marginal entropy estimates.
\State Return $\hH = \hat{S} - \hat{W}$.
\end{algorithmic}
\end{algorithm}

\begin{theorem}\label{thm:det_joint} Algorithm \ref{alg:fastmst} is $\alpha$-locally differentially private and has $O(1)$ communication complexity. Assume that each $X_i$ has constant support size. The expected number of samples requested by the server is $O\left( \frac{d \log \frac{d}{\delta}}{\alpha^2 \epsilon^5}\right)$. 
Let $\hH$ be the entropy estimate output by the algorithm. 
If $X$ is tree-structured, then $|\hH - H(X)| \le \eps d$ with probability $1 - \delta$.
\end{theorem}

\subsection{Our Algorithm for Chain-structured Joint Distributions}

Verma and Pearl~\cite{VePe90} observed that if $X$ is chain-structured with chain $T$ then for any triplet $(X_i$, $X_j, X_k)$, if $X_k$ is on the unique path in $T$ between $X_i$ and $X_j$, then $I( X_i; X_j \vert X_k ) = 0$. 
Thus, for any triplet $(X_i, X_j, X_k)$ in $T$, at least one of $I(X_i; X_j \vert X_k)$, $I(X_i; X_k \vert X_j)$, or $I(X_j; X_k \vert X_i)$ has to be zero.  
This observation alone is not enough to recover the chain, since the conditional mutual information $I( X_i; X_j \vert X_k )$ can also be zero for $X_i, X_j$ and $X_k$ when $X_k$ is not on the path between $X_i$ and $X_j$ in the chain $T$. 
Nevertheless, under the mild assumption that the mutual information $I(X_i, X_j)$ between every pair of variables is distinct, we can recover the chain $T$ by estimating the conditional mutual information of triplets of variables.

Our algorithm is similar to sorting algorithms such as \emph{mergesort} \cite{katajainen1997meticulous}, which require $O(d \log_2 d)$ pairwise comparisons over $d$ items. While we cannot compare pairs explicitly like in a sorting problem, 
for any triplet $(X_i, X_j, X_k)$, we can use their conditional mutual information estimators to locally decide which ``item'' is between the other two: i.e.,  $X_i \leftrightarrow X_j \leftrightarrow X_k$, $X_i \leftrightarrow X_k \leftrightarrow X_j$ or $X_k \leftrightarrow X_i \leftrightarrow X_j$ in the chain $T$. This suggests our Algorithm \ref{alg:markov_chain}, which inserts the variables in a chain one by one in a sequential manner. Algorithm \ref{alg:markov_chain} calls Algorithm \ref{alg:ternary} as a subroutine that seeks to find the position where to insert.

\begin{theorem} \label{thm:markov_chain}
Algorithm \ref{alg:markov_chain} is $\alpha$-locally differentially private and has $O(1)$ communication complexity. 
The number of samples requested by the server is $O\left(\frac{d\log \frac{d}{\delta}}{\alpha^2 \epsilon^2}\right)$. 
Let $\hH$ be the entropy estimate output by the algorithm. 
If $X$ is chain-structured and $\vert I(X_i; X_j) - I(X_j; X_k) \vert \ge \epsilon$, then $|\hH - H(X)| \le \eps d$ with probability $1 - \delta$.
\end{theorem}

\subsection{Our Algorithm for Star-structured Joint Distributions}

If \(X\) is star-structured then recovering the star \(T\) is a matter of identifying its center, which can be done by computing the mutual information between only \(\tilde{\mathcal{O}}(d)\) pairs of variables. The algorithm picks a random marginal \(X_i\) and takes a ``Prim's  step'' \cite{prim1957shortest},  i.e., chooses the neighboring node (say \(X_k\)) that has the largest mutual information with \(X_i\). Assuming that the mutual information $I(X_i, X_j)$ between every pair of variables is distinct, the edge between \(X_i\) and \(X_k\) is in the maximal spanning tree. Next, the algorithm estimates \(\sum_{j\neq i}I(X_i,X_j)\) and \(\sum_{j\neq k}I(X_k,X_j)\) to decide whether \(X_i\) or \(X_k\) is the center node of the star. Algorithm~\ref{alg:star} presents the procedure, and Theorem~\ref{thm:star} gives its sample complexity.

\begin{theorem} \label{thm:star} Algorithm \ref{alg:star} is $\alpha$-locally differentially private and has $O(1)$ communication complexity. 
The number of samples requested by the server is $O\left(\frac{d\log \frac{d}{\delta}}{\alpha^2 \epsilon^2}\right)$. Let $\hH$ be the entropy estimate output by the algorithm. If $X$ is star-structured and $\vert I(X_i; X_j) - I(X_j; X_{k}) \vert \ge \epsilon$, then $|\hH - H(X)| \le \eps d$ with probability \mbox{$1 - \delta$}.
\end{theorem}

\begin{algorithm}[!ht]
\caption{Shannon entropy estimation for chain-structured distribution}\label{alg:markov_chain}
\begin{algorithmic}[1]
    \State $S = [d]$, $C = \emptyset$, pick an arbitrary $i,j,k \in S$ and set $S=S \setminus \{i,j,k \}$.
    \State Server computes $(\alpha, \eps, \delta)$-good estimates $\hat{I}(X_i; X_j \given X_k)$, $\hat{I}(X_i; X_k \given X_j)$ and $\hat{I}(X_k; X_j \given X_i)$.
    \If{$\;\hat{I}(X_i; X_j \given X_k)>\epsilon\;$} 
        ~ $\bx_1 = (i,k,j)$ 
    \ElsIf{$\;\hat{I}(X_i; X_k \given X_j)>\epsilon\;$} 
        ~$\bx_1 = (i,j,k)$
    \ElsIf{$\;\hat{I}(X_k; X_j \given X_i)>\epsilon\;$}
        ~$\bx_1 = (j,i,k)$
    \EndIf 
    \For{$i \in (1,\dots, d-3)$}
        \State Pick item $j$ from $S$ and set $S=S\setminus \{j \}$, $r=x_{i,1}$ and $p=x_{i,i+2}$
        \State Server computes $(\alpha, \eps, \delta)$-good estimates $\hat{I}(X_j; X_p \given X_o)$, $\hat{I}(X_r; X_j \given X_p)$.
        \If{$\;\hat{I}(X_j; X_p \given X_r)>\epsilon\;$} 
            ~ $\bx_{i+1} = (j, \bx_i)$ \Comment{Attach $X_j$ to the head of the chain}
        \ElsIf{$\;\hat{I}(X_o; X_j \given X_p)>\epsilon\;$} 
            ~ $\bx_{i+1} = (\bx_i, j)$ \Comment{Attach $X_j$ to the tail of the chain}
        \Else \Comment{Insert $X_j$ into the chain defined by $\bx_i$}
            \State $\ell =\Algo{TernarySearch} (\bx_i, 1,i+2, j)$ \Comment{Defined in Algorithm \ref{alg:ternary}}
            \State $\bx_{i+1} = (\bx_i[1,\dots, \ell], j,  \bx_i[\ell+1,\dots, i+2] )$
        \EndIf         
    \EndFor
    \State Create chain $T$ according to the order defined by $\bx_{d-2}$.
    \State Server computes $(\alpha, \eps, \delta)$-good estimate of each term in Eq.~\eqref{eq:clmain} using $T$ and returns $\hH$.

\end{algorithmic}
\end{algorithm}

\begin{algorithm}[h!]
\caption{$\Algo{TernarySearch} (\bx, \ell_l,\ell_h, j)$}\label{alg:ternary}
\begin{algorithmic}[1]
\If{$\ell_l = \ell_h - 1$} {\bf return } $\ell_l$
\EndIf
\State Pick the median element $k = \lceil (\ell_h +\ell_l)\rceil$, and set $i=x_{\ell_l}$ and $o=x_{\ell_h}$
\State Server computes $(\alpha, \eps, \delta)$-good estimate $\hat{I}(X_i; X_k \given X_j)$.
\If{$\;\hat{I}(X_i; X_k \given X_j)>\epsilon\;$} 
    ~ {\bf return } $\Algo{TernarySearch} (\bx, i, k, j)$
\Else
    ~{\bf return } $\Algo{TernarySearch} (\bx, k, o, j)$
\EndIf
\end{algorithmic}
\end{algorithm}

\begin{algorithm}[h!]
\caption{Shannon entropy estimation for star-structured distribution}\label{alg:star}
\begin{algorithmic}[1]
    \State Pick $i \in [d]$ uniformly at random. 
    \State Server computes $(\alpha, \eps, \delta)$-good estimate $\hat{I}(X_i, X_j)$ for all $j \in [d] \setminus \{i\}$.
    \State Find $k = \argmax_{j\in [d]\setminus \{ i \}} \hat{I}(X_i, X_j)$
    \State Server computes $(\alpha, \eps, \delta)$-good estimate $\hat{I}(X_k, X_j)$.
    \If{$\;\sum_{j} \hat{I}(X_i, X_j) > \sum_{j} \hat{I}(X_k, X_j)\;$} let $T$ be a star with $X_i$ as center
    \Else $\;$ let $T$ be a star with $X_k$ as the center.
    \EndIf
    \State Server computes $(\alpha, \eps, \delta)$-good estimate of each term in Eq.~\eqref{eq:clmain} using $T$ and returns $\hH$.
\end{algorithmic}
\end{algorithm}

\subsection{Lower Bounds}

We prove sample complexity lower bounds for estimating the Shannon entropy of a tree-structured joint distribution from pair observations. Our first lower bound focuses on the non-interactive case, when the algorithm must select all the pairs in advance. The second claim is more general, and holds for all sequentially interactive algorithms.

\begin{theorem}\label{thm:det_lower}
For any non-interactive algorithm that uses $o(d^2)$ pair observations to estimate Shannon entropy, there exists a tree-structured distribution over $\{ 0,1 \}^d$ such that the error of the algorithm is $\Omega(d)$ with constant probability.
\end{theorem}
\begin{theorem}\label{thm:adaptive_joint}
For any $\epsilon > 0$ and for any sequentially interactive algorithm that uses $o(d/\epsilon)$ pair observations to estimate  Shannon entropy, there exists a tree-structured distribution on $\{0,1 \}^d$ such that the error of the algorithm is $\Omega(\epsilon\cdot d)$ with constant probability.
\end{theorem}
The lower bound given in Theorem \ref{thm:det_lower} is based on Tur\'an's theorem~\cite{Bollobas1998Modern}, which we use to show that for any algorithm with sub-quadratic sample complexity and for any constant $C\in (0,1)$, there is a graph with $d$ nodes containing a $C\cdot d$-clique -- when $d$ is large enough -- such that the algorithm does not observe any edge of that clique. This implies that the additive error of the algorithm is linear in $d$. The lower bound for sequentially interactive algorithms in Theorem  \ref{thm:adaptive_joint} is based on a information theoretical approach. Interestingly, our construction of problem instances for which we applied Le Cam's theorem is fairly simple, since it contains $d$ independent random variables in every case. Nevertheless, this lower bound shows that Algorithm \ref{alg:fastmst} is optimal in $d$.

\subsection{Comparison to Prior Work}

To the best of our knowledge, the Chow-Liu algorithm is the only published method for estimating the entropy of a distribution that takes advantage of its tree structure. Since the algorithm is non-interactive, the lower bound in Theorem \ref{thm:det_lower} shows that our algorithms have provably better sample complexity when the number of variables $d$ is large (note that the dependence on $d$ in each of Theorems \ref{thm:det_joint}, \ref{thm:markov_chain} and \ref{thm:star} is sub-quadratic). The Chow-Liu algorithm can also be used to estimate the distribution itself, not just its entropy, and it has recently been shown \cite{bhattacharyya2021near, daskalakis2020tree} that the algorithm has optimal sample complexity when given full observations (i.e., samples of the entire vector $(X_1, \ldots, X_d)$ and not just pairs or triplets of the variables). Thus the Chow-Liu algorithm is optimal for estimating a tree-structured distribution, but suboptimal for estimating the \emph{entropy} of a tree-structured distribution. The root cause of this difference appears to be the fact that it is significantly easier to estimate the weight of the maximum spanning tree than finding the tree itself.

\section{Estimating Gini and Collision Entropies}
\label{sec:ginicollision}

In this section, we describe a non-iterative protocol (Algorithm \ref{alg:ginicollision}) that estimates both the Gini and collision entropies of a discrete random variable $X$ while observing only $b$ bits per sample from its distribution, and does not require any extra assumption on the distribution of \(X\) (such as assuming it is tree-structured). First, the server partitions all users into pairs (assume for simplicity that the number of users is even). The server then distributes a $b$-bit hash function to each user, along with a distinct salt to each user pair. Each user then hashes their sample along with their salt, and returns the hash value to the server. The server computes entropy estimates based on the number of observed hash collisions across all pairs. In Algorithm \ref{alg:ginicollision}, we let $\inner{x, y}$ denote a binary string that encodes $x$, followed by a delimiter, and by $y$.

\begin{algorithm}[!ht]
\caption{Gini and collision entropy estimation\label{alg:ginicollision}}
\begin{algorithmic}[1]
\Statex
\State Each user $i \in [n]$ draws a sample $x_i$ independently from the distribution of $X$.
\State Server partitions the $n$ users into $\frac{n}{2}$ disjoint pairs.
\State Let $q_i \in \left[\frac{n}{2}\right]$ be the index of the pair containing user $i$.
\State Server transmits $q_i$ and hash function $h : \{0, 1\}^* \mapsto \{0, 1\}^b$ to each user $i$.
\State Each user $i$ generates a $b$-bit hash value $v_i = h(\langle q_i, x_i \rangle)$ for their sample $x_i$.
\State Each user $i$ lets $\hv_i = v_i$ with probability $\lambda = \frac{e^\alpha - 1}{2^b + e^\alpha - 1}$ and else draws $\hv_i$ uniformly from $[2^b]$. \label{alg:ginicollision:line_rr}
\State Server receives $\hv_i$ from each user $i$. 
\State If pair $q$ contains users $i$ and $j$ then let $c_q = \mathbf{1}[\hv_i = \hv_j]$ indicate whether a hash collision was observed for pair $q$.
\State Server computes $\bac = \frac{2}{n} \sum_q c_q$.
\State Server outputs $\hG = \frac{2^b\bac - 1}{\lambda^2 (2^b - 1)}$ and $\hC = -\log \left(1 - \hG\right)$.
\end{algorithmic}
\end{algorithm}

Algorithm \ref{alg:ginicollision} is based on the observation that if $X$ and $X'$ are independent and identically distributed then the Gini entropy is equal to $1 - \Pr[X = X']$ and the collision entropy is equal to $- \log \Pr[X = X']$.  If the server observed each sample directly then it could estimate $\Pr[X = X']$ using the collision frequency, i.e., the fraction of sample pairs $(x_i, x_j)$ such that $x_i = x_j$. However, the server only observes a $b$-bit hash of each sample. Among sample pairs in which there is a true collision, all of them also produce a hash collision. Among samples pairs in which there is not a true collision, about a $\frac{1}{2^b}$ fraction of them produce a hash collision. Therefore the true collision frequency can be estimated using an appropriately bias-corrected hash collision frequency, and the server uses this estimate to approximate the Gini and collision entropies. 

The analysis of Algorithm \ref{alg:ginicollision} is given in Theorem \ref{thm:ginicollision} below. As is customary, for the analysis we assume that the hash function $h$ is constructed by assigning each element of its domain to a uniform random element of its range \cite{bellare1993random}. See the Appendix~\ref{sec:ginicollisionproof} for the proof of the theorem.

\begin{theorem} \label{thm:ginicollision} Algorithm \ref{alg:ginicollision} is $\alpha$-locally differentially private, has $\tilde{O}(b)$ communication complexity and $\tilde{O}(b)$ space complexity. Let $\hG$ and $\hC$ be the outputs of the algorithm. Let $\alpha, \eps, \delta \in (0, 1)$. If
$
n = \Omega\left(\frac{2^{4b} \max\{1 - G(X), 2^{-b}, \eps\}\log \frac1\delta}{\alpha^4\eps^2}\right)
$,
then $|\hG - G(X)| \le \eps$ with probability at least $1 - \delta$. Also, if $X$ has support size $k$ and
$
n = \Omega\left(\frac{2^{4b} k^2 \log \frac1\delta}{\alpha^4\eps^2}\right)
$,
then $|\hC - C(X)| \le \eps$ with probability at least $1 - \delta$.\end{theorem}

\subsection{Comparison to Prior Work}

Recall that Gini entropy is proportional to the second frequency moment. Local differentially private algorithms for estimating frequency moments were recently studied in \cite{butucea2021locally}. Letting \mbox{$b = 1$} in Algorithm \ref{alg:ginicollision} yields a sample complexity of $\tilde{O}(1 / (\alpha^4 \eps^2))$, which is independent of the distribution's support size, unlike the sample complexity of the non-interactive algorithm for estimating the second frequency moment from \cite{butucea2021locally}. Also our algorithm only uses $1$ bit per sample and $\tilde{O}(1)$ space, while the previous algorithm uses $\Omega(k)$ bits per sample and $\Omega(k)$ space, where $k$ is the support size of the distribution. The authors in \cite{butucea2021locally} asked whether there is a non-interactive algorithm for privately estimating frequency moments with a sample complexity that is independent of the distribution's support size. Here we affirmatively answer this open question for the second frequency moment.

The best known algorithm for estimating collision entropy using $\tilde{O}(1)$ space is due to \cite{TRNG}. The sample complexity of their algorithm is $\tilde{O}\left(k/\eps^2\right)$ and its communication complexity is $O(\log k)$ bits per user. 
Letting $b = 1$ in Algorithm \ref{alg:ginicollision} generalizes the previous algorithm to the private and communication-efficient setting. It was shown in \cite{crouch2016stochastic} that (conditioned on a plausible conjecture) any algorithm that estimates collision entropy to within $O(1)$ error using $O(1)$ space requires $\Omega(k)$ samples. 

\section{Experiments}

In this section we present two sets of experiments to support our theoretical findings. First, we demonstrate that Algorithm \ref{alg:fastmst} is indeed able to estimate the Shannon entropy of tree-structured distributions with linear sample complexity in $d$. Thus it has a superior sample complexity compared to the state-of-the-art non-interactive method~\cite{ChLi68,bhattacharyya2021near}, which has a quadratic sample complexity in $d$. The sample complexity is defined here in terms of number of observations from pairs of marginals. In the second set of experiments, we use our Algorithm \ref{alg:ginicollision} to estimate the collision entropy of discrete distributions, and compare its performance to that of the best-known communication efficient, non-private algorithm for this task, Skorski's algorithm~\cite{TRNG}.

{\bf Estimating Shannon entropy:} To estimate the Shannon entropy of a tree-structured distribution as given by Eq.~\eqref{eq:clmain}, the marginal entropies and the mutual information between certain pairs of marginals has to be estimated. 
The Chow-Liu algorithm estimates the mutual information between all pairs of marginals which results in quadratic sample complexity, whereas Algorithm \ref{alg:fastmst} estimates the mutual information only for a linear fraction of pairs. 
Both algorithms estimate the marginal entropy values by sampling the marginals independently from the mutual information estimations, and they both use the same $\epsilon$ additive error and privacy budget $\alpha$ to estimate the mutual information between pairs of marginals. 
Thus it is fair to compare their performance in terms of number of pairs for which they estimate the mutual information. 
For the Chow-Liu algorithm this is always $d^2$, whereas our Algorithm \ref{alg:fastmst} is randomized, thus we evaluate it over $100$ repetitions and report the average. 

\begin{figure}[ht!]
     \centering
     \begin{subfigure}[b]{0.353\textwidth}
        \centering
        \includegraphics[width=\textwidth]{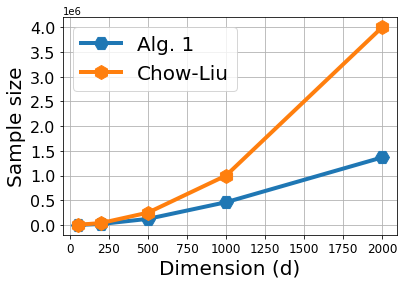}
                \caption{Sample complexity of estimating Shannon entropy of tree-structured distributions (eq.~\eqref{eq:clmain}) for the Chow-Liu algorithm~\cite{ChLi68} and for our Algorithm~\ref{alg:fastmst}.}
        \label{fig:fastmst_alg}
     \end{subfigure}
     \hspace{2cm}
     \begin{subfigure}[b]{0.355\textwidth}
        \centering
        \includegraphics[width=\textwidth]{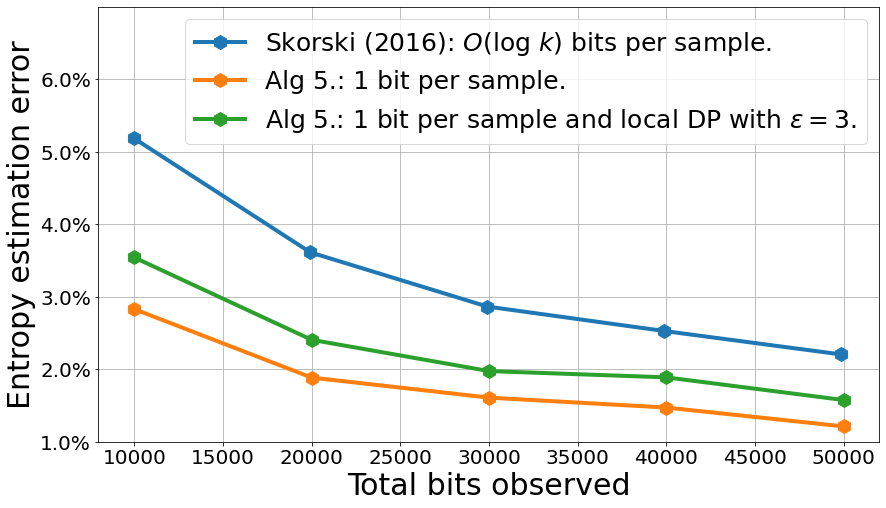}
            \caption{Absolute error in estimating the collision entropy of an exponential distribution with domain size $k=1000$ for the Skorski's algorithm~\cite{TRNG} and for our Algorithm~\ref{alg:ginicollision}. }
        \label{fig:collision_alg}
     \end{subfigure}
     \hfill
\end{figure}

We ran this experiment on random tree-structured joint distributions over $\{0,1\}^d$. 
To create a random tree-structured distribution, we first sample the structure of the distribution by taking the maximum spanning tree of a complete graph with $d$ nodes and edge weights distributed according to a standard normal.
Next, we sample the parameters for each marginal uniformly at random from $[0,1]$, and then we achieve the tree-structure by inducing dependence between pairs of variables while preserving the marginals. 
Specifically, for two marginals $X_i$ and $X_j$ with sampled parameters $p_i$ and $p_j$, we set: 
\mbox{$P(X_i=0,X_j=0)=(1-p_i)\cdot(1- p_j) + r_{ij}$}, 
\mbox{$P(X_i=1,X_j=1)=p_i\cdot p_j + r_{ij}$}, 
\mbox{$P(X_i=0,X_j=1)=(1-p_i)\cdot p_j - r_{ij}$}, and
\mbox{$P(X_i=1,X_j=0)=p_i\cdot(1- p_j)- r_{ij}$}, 
where $r_{ij}$ is sampled uniformly at random from a range of values such that each probability stays positive.
The results are displayed in Figure \ref{fig:fastmst_alg}.  It is clear that the sample size (\emph{i.e.}, number of mutual information estimate) is close to linear for our algorithm, whereas the Chow-Liu algorithm requires a larger sample size. 

{\bf Estimating collision entropy:} In this set of experiments, we drew samples from a discrete exponential distribution $p_i \propto e^{- i}$ with support size $k=1000$ and estimated the collision entropy using algorithm~\cite{TRNG} (the previously best-known communication-efficient algorithm for this task) and our Algorithm~\ref{alg:ginicollision} (with and without local differential privacy).
Algorithm~\cite{TRNG} requires $O(\log k)$ bits per sample and is not private, while our algorithm only requires $1$ bit per sample and is differentially private. 
The results are displayed in Figure \ref{fig:collision_alg}. The previous algorithm has $5\%$ estimation error after observing $10000$ bits, while our algorithm has less than $3.5\%$ estimation error. Thus our algorithm has lower error for the same communication cost while also being local differentially private.

\section{Conclusion and Future Work}

Estimating entropy is of importance in many practical applications. 
In this paper, we studied three widely used entropy measures: Shannon, Gini and collision entropy.
We described estimation algorithms for each entropy that require minimal communication and satisfy local differential privacy. 
We also validated our theoretical results with simulations. 

Our sequentially interactive algorithm for estimating Shannon entropy of high-dimensional tree-structured distributions observes only two of these dimensions per sample and has a sample complexity $O(d/\epsilon^5)$. 
Our approach relies on the celebrated Chow-Liu approximation~\cite{ChLi68}, providing a substantial improvement on the $\Omega(d^2)$ sample complexity of the original non-interactive Chow-Liu algorithm. We also identified two special cases (viz., when the underlying graphical model of the joint distribution is either a chain or star graph) and provided algorithms with a sample complexity of $\tilde{O}(d\log d /\epsilon^2)$. 

Our algorithm for Gini and collision entropy estimation also improved on the state-of-the-art, either by improving the sample complexity and communication complexity of previous work, or by generalizing the best known algorithm to the private and communication-efficient setting. 

A natural extension of our work on Shannon entropy estimation is to consider higher-order correlations in the Chow-Liu decomposition~\cite{KoSz10}. In contrast to the second-order case which reduces to a maximum spanning tree problem, discovering the underlying structure of the joint distribution is already computationally challenging. However, efficiently estimating the entropy of the resulting distribution  might still be possible.

\bibliography{arxiv}

\begin{thebibliography}{10}

\bibitem{pmlr-v89-acharya19b}
Jayadev Acharya, Clement Canonne, Cody Freitag, and Himanshu Tyagi.
\newblock Test without trust: Optimal locally private distribution testing.
\newblock In {\em Proceedings of the Twenty-Second International Conference on
  Artificial Intelligence and Statistics}, volume~89 of {\em Proceedings of
  Machine Learning Research}, pages 2067--2076. PMLR, 2019.

\bibitem{pmlr-v99-acharya19a}
Jayadev Acharya, Cl{\'{e}}ment~L Canonne, and Himanshu Tyagi.
\newblock Inference under information constraints: Lower bounds from chi-square
  contraction.
\newblock In {\em Proceedings of the Thirty-Second Conference on Learning
  Theory}, volume~99 of {\em Proceedings of Machine Learning Research}, pages
  3--17. PMLR, 2019.

\bibitem{acharya2018inspectre}
Jayadev Acharya, Gautam Kamath, Ziteng Sun, and Huanyu Zhang.
\newblock Inspectre: Privately estimating the unseen.
\newblock In {\em International Conference on Machine Learning}, pages 30--39.
  PMLR, 2018.

\bibitem{acharya2019communication}
Jayadev Acharya and Ziteng Sun.
\newblock Communication complexity in locally private distribution estimation
  and heavy hitters.
\newblock In {\em International Conference on Machine Learning}, pages 51--60.
  PMLR, 2019.

\bibitem{acharya2019hadamard}
Jayadev Acharya, Ziteng Sun, and Huanyu Zhang.
\newblock Hadamard response: Estimating distributions privately, efficiently,
  and with little communication.
\newblock In {\em The 22nd International Conference on Artificial Intelligence
  and Statistics}, pages 1120--1129. PMLR, 2019.

\bibitem{bansal2003sublinear}
Vikas Bansal.
\newblock Sublinear-time algorithms for estimating the weight of minimum
  spanning trees.
\newblock {\em Unpublished manuscript}, 109, 2003.

\bibitem{bellare1993random}
Mihir Bellare and Phillip Rogaway.
\newblock Random oracles are practical: A paradigm for designing efficient
  protocols.
\newblock In {\em Proceedings of the 1st ACM Conference on Computer and
  Communications Security}, pages 62--73, 1993.

\bibitem{bhattacharyya2021near}
Arnab Bhattacharyya, Sutanu Gayen, Eric Price, and NV~Vinodchandran.
\newblock Near-optimal learning of tree-structured distributions by
  \uppercase{C}how-\uppercase{L}iu.
\newblock In {\em Proceedings of the 53rd Annual ACM SIGACT Symposium on Theory
  of Computing}, pages 147--160, 2021.

\bibitem{Bollobas1998Modern}
Béla Bollobás.
\newblock {\em Modern Graph Theory}.
\newblock Graduate Texts in Mathematics 184. 1998.

\bibitem{n6}
Tai-Danae Bradley.
\newblock Entropy as a topological operad derivation.
\newblock {\em Entropy}, 23(9), 2021.

\bibitem{butucea2021locally}
Cristina Butucea and Yann Issartel.
\newblock Locally differentially private estimation of functionals of discrete
  distributions.
\newblock In M.~Ranzato, A.~Beygelzimer, Y.~Dauphin, P.S. Liang, and J.~Wortman
  Vaughan, editors, {\em Advances in Neural Information Processing Systems},
  volume~34, pages 24753--24764. Curran Associates, Inc., 2021.

\bibitem{chazelle2001approximating}
Bernard Chazelle, Ronitt Rubinfeld, and Luca Trevisan.
\newblock Approximating the minimum spanning tree weight in sublinear time.
\newblock In {\em International Colloquium on Automata, Languages, and
  Programming}, pages 190--200. Springer, 2001.

\bibitem{ChLi68}
C.~Chow and C.~Liu.
\newblock Approximating discrete probability distributions with dependence
  trees.
\newblock {\em IEEE Transactions on Information Theory}, 14(3):462--467, 1968.

\bibitem{Cover2006}
Thomas~M. Cover and Joy~A. Thomas.
\newblock {\em Elements of Information Theory (Wiley Series in
  Telecommunications and Signal Processing)}.
\newblock Wiley-Interscience, New York, NY, USA, 2006.

\bibitem{crouch2016stochastic}
Michael Crouch, Andrew McGregor, Gregory Valiant, and David~P Woodruff.
\newblock Stochastic streams: Sample complexity vs. space complexity.
\newblock In {\em 24th Annual European Symposium on Algorithms (ESA 2016)}.
  Schloss Dagstuhl-Leibniz-Zentrum fuer Informatik, 2016.

\bibitem{czumaj2004estimating}
Artur Czumaj and Christian Sohler.
\newblock Estimating the weight of metric minimum spanning trees in
  sublinear-time.
\newblock In {\em Proceedings of the thirty-sixth annual ACM symposium on
  Theory of computing}, pages 175--183, 2004.

\bibitem{daskalakis2020tree}
Constantinos Daskalakis and Qinxuan Pan.
\newblock Tree-structured \uppercase{I}sing models can be learned efficiently.
\newblock {\em arXiv e-prints}, pages arXiv--2010, 2020.

\bibitem{duchi2019lower}
John Duchi and Ryan Rogers.
\newblock Lower bounds for locally private estimation via communication
  complexity.
\newblock In {\em Conference on Learning Theory}, pages 1161--1191. PMLR, 2019.

\bibitem{dwork2014algorithmic}
Cynthia Dwork, Aaron Roth, et~al.
\newblock The algorithmic foundations of differential privacy.
\newblock {\em Found. Trends Theor. Comput. Sci.}, 9(3-4):211--407, 2014.

\bibitem{fukuchi2017minimax}
Kazuto Fukuchi and Jun Sakuma.
\newblock Minimax optimal estimators for additive scalar functionals of
  discrete distributions.
\newblock In {\em 2017 IEEE International Symposium on Information Theory
  (ISIT)}, pages 2103--2107. IEEE, 2017.

\bibitem{n8}
C.~Gini.
\newblock {\em Variabilit\'a e Mutuabilit\'a. Contributo allo Studio delle
  Distribuzioni e delle Relazioni Statistiche}.
\newblock Bologna: C. Cuppini., 1912.

\bibitem{han2018geometric}
Yanjun Han, Ayfer {\"O}zg{\"u}r, and Tsachy Weissman.
\newblock Geometric lower bounds for distributed parameter estimation under
  communication constraints.
\newblock In {\em Conference On Learning Theory}, pages 3163--3188. PMLR, 2018.

\bibitem{Herfindahl1950ConcentrationIT}
Oc~Herfindahl.
\newblock Concentration in the \uppercase{US} steel industry.
\newblock 1950.

\bibitem{joseph2019role}
Matthew Joseph, Jieming Mao, Seth Neel, and Aaron Roth.
\newblock The role of interactivity in local differential privacy.
\newblock In {\em 2019 IEEE 60th Annual Symposium on Foundations of Computer
  Science (FOCS)}, pages 94--105. IEEE, 2019.

\bibitem{JosephMN019}
Matthew Joseph, Jieming Mao, Seth Neel, and Aaron Roth.
\newblock The role of interactivity in local differential privacy.
\newblock In David Zuckerman, editor, {\em 60th {IEEE} Annual Symposium on
  Foundations of Computer Science, {FOCS} 2019, Baltimore, Maryland, USA,
  November 9-12, 2019}, pages 94--105. {IEEE} Computer Society, 2019.

\bibitem{katajainen1997meticulous}
Jyrki Katajainen and Jesper~Larsson Tr{\"a}ff.
\newblock A meticulous analysis of mergesort programs.
\newblock In {\em Italian Conference on Algorithms and Complexity}, pages
  217--228. Springer, 1997.

\bibitem{KoSz10}
Edith Kov{\'a}cs and Tam{\'a}s Sz{\'a}ntai.
\newblock {\em On the Approximation of a Discrete Multivariate Probability
  Distribution Using the New Concept of t-Cherry Junction Tree}, pages 39--56.
\newblock Springer Berlin Heidelberg, Berlin, Heidelberg, 2010.

\bibitem{LaaksoTaagepera}
Markku Laakso and Rein Taagepera.
\newblock “\uppercase{E}ffective” number of parties: A measure with
  application to west europe.
\newblock {\em Comparative Political Studies}, 12(1):3--27, 1979.

\bibitem{n1}
Tom Leinster.
\newblock {\em Entropy and Diversity: The Axiomatic Approach}.
\newblock Cambridge University Press, 2021.

\bibitem{motahari2013information}
Abolfazl~S Motahari, Guy Bresler, and NC~David.
\newblock Information theory of \uppercase{DNA} shotgun sequencing.
\newblock {\em IEEE Transactions on Information Theory}, 59(10):6273--6289,
  2013.

\bibitem{prim1957shortest}
Robert~Clay Prim.
\newblock Shortest connection networks and some generalizations.
\newblock {\em The Bell System Technical Journal}, 36(6):1389--1401, 1957.

\bibitem{raileanu2004theoretical}
Laura~Elena Raileanu and Kilian Stoffel.
\newblock Theoretical comparison between the {G}ini index and information gain
  criteria.
\newblock {\em Annals of Mathematics and Artificial Intelligence},
  41(1):77--93, 2004.

\bibitem{n4}
Alfr\'ed R\'enyi.
\newblock On measures of information and entropy.
\newblock In {\em Proceedings of the fourth Berkeley Symposium on Mathematics,
  Statistics and Probability}, pages 547--561, 1960.

\bibitem{n5}
C.E. Shannon.
\newblock A mathematical theory of communication.
\newblock {\em Bell Syst. Tech. J.}, 27:379--423, 1948.

\bibitem{simpson1949measurement}
Edward~H Simpson.
\newblock Measurement of diversity.
\newblock {\em nature}, 163(4148):688--688, 1949.

\bibitem{n7}
Yakov~G. Sinai.
\newblock On the notion of entropy of a dynamical system.
\newblock 2010.

\bibitem{TRNG}
Maciej Skorski.
\newblock Evaluating entropy for true random number generators: Efficient,
  robust and provably secure.
\newblock In {\em International Conference on Information Security and
  Cryptology}, pages 526--541, 03 2017.

\bibitem{suresh2017distributed}
Ananda~Theertha Suresh, X~Yu Felix, Sanjiv Kumar, and H~Brendan McMahan.
\newblock Distributed mean estimation with limited communication.
\newblock In {\em International Conference on Machine Learning}, pages
  3329--3337. PMLR, 2017.

\bibitem{n3}
Constantino Tsallis.
\newblock {\em Introduction to Nonextensive Statistical Mechanics}.
\newblock Springer New York, NY, Greece, 2009.

\bibitem{VePe90}
Thomas Verma and Judea Pearl.
\newblock Equivalence and synthesis of causal models.
\newblock In {\em Proceedings of the Sixth Annual Conference on Uncertainty in
  Artificial Intelligence}, UAI '90, page 255–270, USA, 1990. Elsevier
  Science Inc.

\bibitem{zhang2013information}
Yuchen Zhang, John~C Duchi, Michael~I Jordan, and Martin~J Wainwright.
\newblock Information-theoretic lower bounds for distributed statistical
  estimation with communication constraints.
\newblock In {\em NIPS}, pages 2328--2336. Citeseer, 2013.

\end{thebibliography}
\bibliographystyle{plain}

\newpage

\appendix



\newpage
\section{Proof of Theorem \ref{thm:privateshannon}}

\begin{proof}
\cite{acharya2019communication,acharya2019hadamard} showed that any discrete distribution can be learnt in total variation distance based on $O(c^2\log1/\delta / (\epsilon^2\alpha^2)$ when $\alpha<1$. This result can be plug-in into Theorem 17.3.3 of \cite{Cover2006}.
\end{proof}

\section{Proof of Theorem \ref{thm:det_joint}}

Existing algorithms for estimating the weight of the maximum spanning tree in sublinear time assume that each edge weight can be computed in $O(1)$ time. Our main idea is to combine one of these algorithms with Theorem \ref{thm:privateshannon}, which gives the number of samples needed to privately estimate an edge weight (i.e., the mutual information between two variables).

The first sublinear algorithm for MST weight estimation is due to \cite{chazelle2001approximating}. Our algorithm and analysis follows the exposition in \cite{bansal2003sublinear}, which is itself based on \cite{czumaj2004estimating}. The main technical complication we must overcome is that the results from \cite{chazelle2001approximating} and \cite{bansal2003sublinear} assume that the edge weights have a bounded ratio, while we instead discretize the edge weights with resolution $\eps$.

Let $G$ be a graph on $d$ nodes such that the edge weight between nodes $i$ and $j$ is $I(X_i; X_j)$. Let $\tG$ be a graph on $d$ nodes with edge weights belonging to the set
\[
\{\eps, 2\eps, \ldots, M\eps\}
\]
and such that every edge weight in $\tG$ differs from the corresponding edge weight in $G$ by at most $\eps$. Note that this is possible because the edge weights in $G$ are guaranteed to be in the interval $[0, M\eps]$. Let $w(G)$ be the weight of a maximum spanning tree of $G$. Clearly we have
\begin{equation}
    |w(G) - w(\tG)| \le \eps d. \label{eq:mst:one}
\end{equation}

For each $m \in \{1, \ldots, M\}$ let $\tG_m$ be the subgraph of $\tG$ containing the same nodes as $\tG$ and only those edges of $\tG$ whose weight is at least $m\epsilon$. Thus $\tG_1 = \tG$.

Let $\CC_m$ be the set of connected components in $\tG_m$. Also let
\begin{align*}
    \CC^-_m &= \left\{C \in \CC_m : |C| \le \frac2\eps\right\}\\
    \CC^+_m &= \left\{C \in \CC_m : |C| > \frac2\eps\right\}
\end{align*}
be the sets of connected components in $\tG_m$ with size at most $\frac2\eps$ and greater than $\frac2\eps$, respectively. Let $\eta_m = |\CC_m|$, $\eta^-_m = |\CC^-_m|$ and $\eta^+_m = |\CC^+_m|$. Clearly
\begin{equation}
    \eta^+_m \le \frac{\eps d}{2} \label{eq:mst:two}
\end{equation}
since $\tG_m$ has $d$ nodes. Now consider the iteration of Algorithm \ref{alg:fastmst} that sets the value of $\gamma_{mr}$. Let $i^*$ and $Z$ be the random variables from this iteration. Recall that an $\left(\alpha, \frac{\eps}{2}, \frac{\delta}{d^2}\right)$-good estimate is computed for the weight of each edge. Therefore, with probability $1 - \delta$, this iteration performs a breadth-first search on the graph $\tG_m$, and sets $\gamma_{mr} = 1$ if and only if the component in $\tG_m$ containing $i^*$ has size at most $\min\left\{\frac1\eps, Z\right\}$. Thus we have
\begin{align*}
\E[\gamma_{mr}] &= \sum_{C \in \CC_m} \Pr[i^* \in C] \Pr\left[|C| \le \min\left\{\frac2\eps, Z\right\}\right]\\
&= \sum_{C \in \CC^-_m} \Pr[i^* \in C] \Pr\left[|C| \le Z\right]\\
&= \sum_{C \in \CC^-_m} \frac{|C|}{d} \frac{1}{|C|}\\
&= \frac{\eta_m^-}{d}.
\end{align*}
Since $i^*$ and $Z$ are drawn independently in each iteration, we have by Hoeffding's inequality with probability $1 - \delta$
\begin{equation}
    \left\lvert \heta_m - \eta^-_m\right\rvert = \left\lvert \frac{d}{r} \sum_{r = 1}^R \gamma_{mr} - \eta^-_m\right \rvert \le \eps d. \label{eq:mst:three}
\end{equation}

Next, we will relate the weight of the maximum spanning tree of $\tG$ to the number of connected components in various subgraphs of $\tG$. For each $m \in \{1, \ldots, M\}$ let $\beta_m$ be the number of edges with weight $m\epsilon$ in a maximum spanning tree of $\tG$. We have
\[
\sum_{m < \ell} \beta_m = \eta_\ell - 1
\]
for any $\ell \in \{1, \ldots, M\}$, where the empty sum is defined to be zero. This equality can be established by considering Kruskal's greedy algorithm for constructing a maximum spanning tree, which adds edges in decreasing order of weight as long as they do not induce a cycle. Since $\tG_\ell$ has $\eta_\ell$ connected components, and all the edges in $\tG_\ell$ are heavier than all the edges not in $\tG_\ell$, the greedy algorithm must first connect the vertices within each component of $\tG_\ell$ and then use exactly $\eta_\ell - 1$ edges not in $\tG_\ell$ to connect the components to each other. Thus we have
\begin{align}
    w(\tG) &= \sum_{m=1}^M m\epsilon\beta_m \notag\\
    &= \epsilon \left(\sum_{m \ge 1} \beta_m + \sum_{m \ge 2} \beta_m + \cdots + \sum_{m \ge M} \beta_m\right) \notag\\
    &= \epsilon \left(d - 1 - \sum_{m < 1} \beta_m + d - 1 - \sum_{m < 2} \beta_m + \cdots + d - 1 - \sum_{m < M} \beta_m\right) \notag\\
    &= \epsilon \left(M(d - 1) - \sum_{m=1}^M (\eta_m - 1)\right) \notag\\
    &= \epsilon Md - \epsilon\sum_{m=1}^M \eta_m \label{eq:mst:four}
\end{align}
Putting everything together, and using the definition of $\hat{W}$ from Algorithm \ref{alg:fastmst}, we have with probability $1 - \delta$
\begin{align*}
\left\lvert \hat{W} - w(G) \right\rvert &= \left \lvert \eps Md - \eps \sum_{m=1}^M \heta_m - w(G) \right\rvert\\
&\le \left \lvert \eps Md - \eps \sum_{m=1}^M \eta_m - w(G) \right\rvert + \eps \sum_{m=1}^M \left\lvert \eta^-_m - \heta_m\right\rvert + \eps \sum_{m=1}^M \left\lvert \eta^-_m - \eta_m\right\rvert\\
&= \left \lvert \eps Md - \eps \sum_{m=1}^M \eta_m - w(G) \right\rvert + \eps \sum_{m=1}^M \left\lvert \eta^-_m - \heta_m\right\rvert + \eps \sum_{m=1}^M \left\lvert \eta^+_m\right\rvert\\
&= \left \lvert w(\tG) - w(G) \right\rvert + \eps \sum_{m=1}^M \left\lvert \eta^-_m - \heta_m\right\rvert + \eps \sum_{m=1}^M \left\lvert \eta^+_m\right\rvert & \because \textrm{Eq.~\eqref{eq:mst:four}}\\
&\le \left \lvert w(\tG) - w(G) \right\rvert + \eps^2 M d + \eps \sum_{m=1}^M \left\lvert \eta^+_m\right\rvert & \because \textrm{Eq.~\eqref{eq:mst:three}}\\
&\le \left \lvert w(\tG) - w(G) \right\rvert + \eps^2 M d + \frac{\eps^2 M d}{2} & \because \textrm{Eq.~\eqref{eq:mst:two}}\\
&\le \eps d + \eps^2 M d + \frac{\eps^2 M d}{2} & \because \textrm{Eq.~\eqref{eq:mst:one}}\\
&= O(\eps d).
\end{align*}
Finally, recall that an $\left(\alpha, \eps, \frac{\delta}{d}\right)$-good estimate is computed for each marginal entropy. Therefore with probability $1 - \delta$ we have
\[
\left \lvert\hat{S} - \sum_i H(X_i)\right\rvert \le \eps d
\]
and thus by Eq.~\eqref{eq:clmain} we have 
\[
|\hat{H} - H(X)| = \left \lvert \hat{S} - \hat{W} - \left(\sum_i H(X_i) - w(G)\right)\right \rvert = O(\eps d).
\]

Algorithm \ref{alg:fastmst} has $\tilde{O}\left(\frac{1}{\eps^3}\right)$ iterations, in each iteration estimates the weights of
\[
d \sum_{z \le \frac2\eps} z \Pr[Z = z] = d \sum_{z \le \frac2\eps} \Pr[Z \ge z] = O\left(d \log\frac1\eps\right)
\]
edges on average, and uses $\tilde{O}\left(\frac{1}{\alpha^2\eps^2}\right)$ samples to estimate the weight of each edge.






\section{Proof of Theorem \ref{thm:markov_chain}}
\label{app::markov_chain}

\begin{proof}
We start by recalling a lemma that applies to tree-structured distributions.
\begin{lemma}{\cite{VePe90}} \label{lemma:condint_treedecomp_succ}
Let $\bX = (X_1, \dots, X_d )$ be tree decomposable with tree $T$. Then for any triplets $i$, $j$ and $k$, if $k$ is on the unique path in $T$ between $i$ and $j$, then
\[
I( X_i; X_j \vert X_k ) = 0 \enspace .
\]
\end{lemma}
Next we show that the output Algorithm \ref{alg:markov_chain} is correct with high probability. We make use of the Conditional Mutual Information Tester of \cite{bhattacharyya2021near}. This testing algorithm consists of estimating the CMI using the plug-in estimator and then applying a $\epsilon$ threshold on the estimate, i.e., if the estimate is smaller than $\epsilon$ then accept, otherwise reject. The sample complexity of Conditional Mutual Information Tester is $O\left(\tfrac{\vert \Sigma \vert^3}{\epsilon} \log \frac{d\vert \Sigma \vert}{\delta} \log \frac{\vert \Sigma \vert \log d/\delta}{\epsilon}\right)$ according to Theorem 1.3 of \cite{bhattacharyya2021near}. Thus if we apply this tester with adjusted confidence parameter, i.e., $\delta / d \log_3 d$, then the union bound implies that the output of all tests is correct with probability at least $1-\delta$.

Next, note that for any triplet $X_i,X_j,X_k$ such that $X_i$ is between $X_j$ and $X_k$ in the chain, it holds that
\[
I(X_j; X_{i} ) - I(X_j; X_k ) = \underbrace{I(X_j; X_{i} \vert X_k )}_{> \epsilon } - \underbrace{I(X_j; X_k \vert X_{i} )}_{=0~\text{due to Lemma ~\ref{lemma:condint_treedecomp_succ}}}
\]
as the edges are different with a margin of $\epsilon$. The same argument implies that $I(X_k; X_{i} \vert X_j ) > \epsilon$. Thus, Algorithm \ref{alg:markov_chain} divides the nodes correctly in Line 4-10 which along with the testers' correctness with high probability implies the correctness of the algorithm.

Using the result from \cite{bhattacharyya2021near} that the sample complexity of Conditional Mutual Information Tester is $O\left(\tfrac{\vert \Sigma \vert^3}{\epsilon} \log \frac{d\vert \Sigma \vert}{\delta} \log \frac{\vert \Sigma \vert \log d/\delta}{\epsilon}\right)$, we only need to upper bound the number of tests for deciding whether $I (X_i; X_j \given X_k ) > \epsilon$ that is carried out by Algorithm \ref{alg:markov_chain}. It is indeed $3 \sum_{i=3}^d \log_3 i \in O( d\log_3 d)$ which concludes the proof.
\end{proof}

\section{Proof of Theorem \ref{thm:star}}

\begin{proof}
First note that the Prim step in Line 3 of Algorithm \ref{alg:star} indeed finds an edge that is in the maximum spanning tree due to the assumption $\vert I(X_i; X_j) - I(X_j; X_{k}) \vert \ge \epsilon$. Say this edge is between $X_i$ and $X_k$, what remains to be decided is whether $X_i$ or $X_k$ is the center of the graph.  This can be done by comparing $\sum_{j} \hat{I}(X_i, X_j)$ and $\sum_{j} \hat{I}(X_k, X_j)$. 

So Algorithm \ref{alg:star} estimates $2d$ mutual information, which requires $4d$ marginal and $2d$ pairwise marginal entropy estimation. 
Recall from Theorem \ref{thm:privateshannon} that entropy can estimated  $\alpha$-locally differential private with $\epsilon$ error using $O(c^2 \log\frac1\delta /(\epsilon^2 \alpha^2))$ samples. 
Thus by the union bound the algorithm is correct if the confidence parameter is set to $\delta/6d$, and accordingly has sample complexity   $O(\frac{6dc^2}{\epsilon^2 \alpha^2} \log\frac{6d}{\delta})$.
\end{proof}

\section{Proof of Theorem \ref{thm:det_lower}}
\label{app:det_lower}

\begin{proof}
Assume a deterministic algorithm $\mathcal{A}$ that takes sub-quadratic samples from $\bX$ and estimates $H(\bX)$. 
In addition, assume that its sample complexity is $o( d^{2} )$. 
Thus for any constant $C >0$, there exists $d_0$ such that for any $d > d_0$, the sample complexity of the algorithm is $ < C d^2$.  
This implies that, if $d$ is large enough, the algorithm needs $ \le d^{2-\kappa}$. 
In addition, we can pick $d$ so as $d^{-\kappa} < \kappa$ which implies $d^2 - d^{2-\kappa} > (1-\kappa) d^2$. 
Thus any deterministic algorithm which takes sub-quadratic sample size never observes $(1-\kappa) d^2$ edges for $\forall \kappa $ when $d$ is large enough. 

Let us recall that Tur\'{a}n's theorem
\begin{theorem}[\cite{Bollobas1998Modern}]
Let $G(V,E)$ be a graph with $d$ vertices (i.e, $|V|=d$) that does not contain a $(\ell+1)$-clique as a subgraph. 
Then $G$ has at most 
$\frac{(\ell -1 )d^2}{2\ell}$ 
edges. 
\end{theorem}
In addition, recall that the Tur\'{a}n's graph $G_T(d,\ell)$~\cite{Bollobas1998Modern} is defined as the unique graph with $d$ nodes that does not contain a $(\ell+1)$-clique and has the maximum possible number of edges which is $\left\lfloor\frac{(\ell-1)d^2}{2\ell} \right\rfloor$.
Let $t(d,\ell)$ denote the number of edges in $G_T(d,\ell)$.

Thus for any $d > 0$ there exists a graph $G = (V, E)$ such that $\vert V \vert =d $ and $\vert E \vert > t(d,k)$ which contains a $(k+1)$-clique. 
This implies that if algorithm $\mathcal{A}$ does not observes at least $t(d,k)$ edges of $G$, 
then $G$ contains a $k+1$-clique whose edges are never observed by algorithm $\mathcal{A}$. 
Now we apply Turan's result with $\ell=d/2$ which implies that
\[
t( d, d/2 ) =\left\lfloor (1/2 - 1/d) d^2 \right\rfloor
\]
Thus for any $\kappa \in (0, 1/2)$ and any algorithm with sample complexity $o(d^2)$, if $d$ is large enough, then there will be a $\Theta(d/2)$-clique for which the algorithm does not observe any edge within this clique.

Finally it is easy to construct two $d/2$-dimensional problem instances, denoted it $S$ and $S'$, with joint entropy that differs by $\Omega(d)$: 
take $d/2$ Bernoulli with parameter $1/2$ and take the copy of the same Bernoulli $d/2$ times. 
The entropy for these two joint distributions is $d/2$ and $1$, respectively. 
This also implies that for any deterministic algorithm we can construct two problem instances which contains $S$ and $S'$ so as they are independent from the rest of the marginals and the algorithm does not observe any sample from them, and hence it cannot achieve $o(d)$ additive error. 
\end{proof}

\section{Proof of Theorem \ref{thm:adaptive_joint}}
\label{app:lower_gen}

\begin{proof}
Let $\hat{\theta}_n = \hat{\theta} ( x_1, \dots, x_n )$ such that $\hat{\theta}_n : (\Sigma^d)^n \mapsto \mathbb{R}$ be an estimator using $n$ samples.

\begin{theorem}{[Le Cam's theorem]}
Let $\mathcal{P}$ be a set of distributions. Then, for any pair of distributions $P_0, P_1 \in \mathcal{P}$, we have
\[
\inf_{\hat{\theta}} \max_{P\in \mathcal{P}} \mathbb{E}_P \left[ d( \hat{\theta}_n (P), \theta(P) )  \right] \ge \frac{d( \theta(P_0), \theta(P_1) )}{8} e^{-n d_{\text{KL}} (P_0, P_1)},
\]
where $\theta ( P)$ is a parameter taking values in a metric space with metric $d$, and $\hat{\theta}_n$ is the estimator of $\theta $ based on $n$ samples.
\end{theorem}

Let us consider two Bernoulli distributions $P_0$ and $P_1$ with parameters $p_0=1/2$ and $p_1=1/2-\epsilon$, where $\epsilon \in (0, 1/2)$. The entropy of random variables $X_0$ and $X_1$ distributed according to $P_0$ and $P_1$ are $H(X_0)=1$ and
\begin{align*}
H(X_1) 
& = -\left(\frac{1}{2} -\epsilon \right) \log_2 \left( \frac{1}{2} -\epsilon \right) -\left( \frac{1}{2} + \epsilon \right) \log_2 \left( \frac{1}{2} + \epsilon \right).
\end{align*}
Thus,
\begin{align*}
\vert H(X_0) - H(X_1)\vert = H(X_0) - H(X_1) &=  
\left(\epsilon + \frac{1}{2}\right) \log_2 ( 1+2\epsilon) - \left(\epsilon - \frac{1}{2}\right) \log_2 (1-2\epsilon) \\
&\geq \left(\epsilon + \frac{1}{2}\right) \frac{2\epsilon}{1+2\epsilon} - \left(\frac{1}{2} - \epsilon\right) 2\epsilon \\
& \ge 2\epsilon^2 
\end{align*}
where we used that $\log(1-2\epsilon)\leq -\epsilon$ for $0<\epsilon<1$ and $\frac{\epsilon}{1+\epsilon}\leq\log(1+\epsilon)$ for $\epsilon>-1$. 
The KL divergence can be upper bounded as
\[
d_{\text{KL}} (P_0, P_1) = -\frac{1}{2}\log_2 (1-4\epsilon^2) \le 2\epsilon^2.
\]
We can now apply the Le Cam's theorem for the set of Bernoulli distributions with metric $d$ being the $\ell_1$-norm as 
\[
\inf_{\hat{\theta}} \max_{P\in \mathcal{P}} \mathbb{E}_P \left[|\hat{\theta}_n (P) - H(P)|\right] \geq \frac{d( \theta(P_0), \theta(P_1) )}{8} e^{-n d_{\text{KL}} (P_0, P_1)} \ge \frac{\epsilon^2}{4} e^{-2n\epsilon^2}
\]
Using this result with $\epsilon' = \sqrt{\epsilon}$, the following sample complexity can be obtained for estimating Shannon entropy.
\begin{corollary}\label{corr:lower_bern}
For any $\hat{\theta}_n$ such that $n\in o(1/\epsilon)$, there exists a Bernoulli distribution $P$ for which 
\[
\mathbb{E}_P \left[ \vert \hat{\theta}_n (P) - H(P) \vert  \right] \ge C\cdot \epsilon, 
\]
with $C>0$ that does not depend on $\epsilon$.
\end{corollary}

First, note that there is some bound on error $r(\delta)$, either lower or upper, that holds with probability $1-\delta$. 
This translate into the bound $r(\delta) +\delta $ on the expected error in a straightforward manner. 
Thus the lower bound presented in Corollary \ref{corr:lower_bern} also implies that there is no high probability estimator for entropy with $o(1/\epsilon)$ sample complexity for discrete distributions. 
This can be used to lower bound of the entropy estimator for joint distribution as follows. 

Let $\mathcal{B} = \left\{\mathbf{b} = ( b_1, \dots , b_d ) : b_j \in \{0, 1\}\right\}$ be the vertices of the $d$ dimensional hypercube,  
and define a set of $d$-dimensional distribution $\mathcal{P}_{\mathbf{b}}$ indexed by the element of $\mathcal{B}$. 
Each $P_\mathbf{b}\in \mathcal{P}$ contains $X_0 \sim \text{Bern}(1/2)$ if $b_i=0$ and $X_1 \sim \text{Bern}(1/2-\epsilon)$ if $b_i=1$, i.e.
\[
P_{\mathbf{b} } = X_{b_1} \oplus \cdots \oplus X_{b_d}
\]
and 
\[
\mathcal{P} = \left\{ P_{\mathbf{b} }: \mathbf{b} \in  \{ 0, 1 \}^d \right\} \enspace .
\]
It is clear that $\mathcal{P}$ is a subset of the tree-structured distributions and each distribution contains $d$ independent Bernoulli random variables, thus
\[
H(P_{\mathbf{b}}) = \sum_{i=1}^d H(X_{b_i})
\]
Therefore any estimator that achieves at most $\epsilon \cdot d$ additive error for $H(P_{\mathbf{b}})$ has to estimate each individual Bernoulli distribution with at most $\epsilon$ error. The sample complexity of any estimator of $H(P_{\mathbf{b}})$ with an additive error $O(\epsilon d)$ is $\Omega( d / \epsilon )$.
\end{proof}




\section{Proof of Theorem \ref{thm:ginicollision}}
\label{sec:ginicollisionproof}

\begin{proof}
Algorithm \ref{alg:ginicollision} clearly has $\tilde{O}(b)$ communication complexity and $\tilde{O}(b)$ space complexity, since it only has to maintain a counter of collisions between $b$-bit hashes. 

To prove that Algorithm 5 is $\alpha$-locally differentially private, note that given $v_i$ the distribution of $\hv_i$ is independent of every other random variable. Also note that each $v_i$ is deterministic function of $h$, $q_i$ and $x_i$. Therefore it suffices to show that for each user $i$ and all $v, v', \hv \in \{1, \ldots, 2^b\}$ we have
\[
\frac{\Pr[\hv_i = \hv ~|~ v_i = v]}{\Pr[\hv_i = \hv ~|~ v_i = v']} \le e^\alpha.
\]
Since user $i$ replaces $v_i$ with a hash value chosen uniformly at random from $\{1, \ldots, 2^b\}$ with probability $1 - \lambda$, we have
\[
\frac{\Pr[\hv_i = \hv ~|~ v_i = v]}{\Pr[\hv_i = \hv ~|~ v_i = v']} \le \frac{\lambda + (1 - \lambda)/2^b}{(1 - \lambda)/2^b} = 1 + \frac{\lambda 2^b}{1 - \lambda} = e^\alpha
\]
where the last equality substitutes $\lambda = \frac{e^\alpha - 1}{2^b + e^\alpha - 1}$.

To prove the sample complexity results, let
\[
\baG(X) = 1 - G(X) = \sum_x \Pr[X = x]^2 = \Pr[X = X']
\]
where $X'$ is independent and identically distributed as $X$. We will calculate the expected value of each $c_q$. Suppose pair $q$ contains samples $x_i$ and $x_j$. Since $q_i = q_j$, we have that if $x_i = x_j$ then $c_q = 1$ with probability $\lambda^2 + \frac{1 - \lambda^2}{2^b}$, and otherwise $c_q = 1$ with probability $\frac{1}{2^b}$. Thus
\begin{align}
\E[c_q] &= \Pr[c_q = 1 ~|~ x_i = x_j] \Pr[x_i = x_j] + \Pr[c_q = 1 ~|~ x_i \neq x_j] \Pr[x_i \neq x_j] \notag\\
&= \left(\lambda^2 + \frac{1 - \lambda^2}{2^b}\right)\Pr[x_i = x_j] + \frac{1}{2^b} \Pr[x_i \neq x_j] \notag\\
&= \left(\lambda^2 + \frac{1 - \lambda^2}{2^b}\right)\Pr[x_i = x_j] + \frac{1}{2^b} (1 - \Pr[x_i = x_j]) \notag\\
&= \lambda^2 \left(1 - \frac{1}{2^b}\right)\Pr[x_i = x_j] + \frac{1}{2^b} \notag\\
&= \lambda^2 \left(1 - \frac{1}{2^b}\right) \baG(X) + \frac{1}{2^b} \label{eq:zero}
\end{align}
where the last line follows because $x_i$ and $x_j$ are independent samples from the distribution of $X$.

Let $m = \frac{n}{2}$. Note that the $c_q$'s are independent random variables because each $c_q$ is defined using a distinct pair of samples and distinct pair index. Also, each $c_q \in \{0, 1\}$. Therefore, by the Chernoff bound, for all $\eps_0 \ge 0$ the average $\bac = \frac{1}{m} \sum_q c_q$ satisfies
\[
\Pr\left[\left\lvert\bac - \E[\bac]\right\rvert \ge \eps_0 \E[\bac]\right] \le 2\exp\left(-\frac{\eps_0^2 m}{2 + \eps_0} \E[\bac]\right).
\]
We proved in Eq.~\eqref{eq:zero} that $\E[\bac] = \lambda^2 \left(1 - \frac{1}{2^b}\right) \baG(X) + \frac{1}{2^b}$. Therefore for all $\eps_0 \ge 0$
\begin{align*}
& \Pr\left[\left\lvert\hG - G(X)\right\rvert \ge \eps_0 \left(\baG(X) + \frac{1}{\lambda^2 (2^b - 1)}\right)\right]\\
=& \Pr\left[\left\lvert\frac{2^b\bac - 1}{\lambda^2(2^b - 1)} - \baG(X)\right\rvert \ge \eps_0 \left(\baG(X) + \frac{1}{\lambda^2 (2^b - 1)}\right)\right] & \because \baG(X) = 1 - G(X)\\
=& \Pr\left[\left\lvert\bac - \frac{1}{2^b} - \lambda^2\left(1 - \frac{1}{2^b}\right)\baG(X)\right\rvert \ge \eps_0 \left(\lambda^2\left(1 - \frac{1}{2^b}\right)\baG(X) + \frac{1}{2^b}\right)\right] & \because \textrm{Multiply by }\frac{\lambda^2(2^b - 1)}{2^b}\\
=& \Pr\left[\left\lvert\bac - \E[\bac]\right \rvert \ge \eps_0 \E[\bac]\right] & \because \textrm{Eq.~\eqref{eq:zero}}\\
\le& 2\exp\left(-\frac{\eps_0^2 m}{2 + \eps_0}\E[\bac]\right) & \because \textrm{Chernoff}\\
=& 2\exp\left(-\frac{\eps_0^2 m}{2 + \eps_0}\left(\lambda^2\left(1 - \frac{1}{2^b}\right) \baG(X) + \frac{1}{2^b}\right)\right)& \because \textrm{Eq.~\eqref{eq:zero}}\\
=& 2\exp\left(-\frac{\eps_0^2 m}{(2 + \eps_0)2^b}\left(\lambda^2\left(2^b - 1\right) \baG(X) + 1\right)\right)
\end{align*}
Recalling that $\rho = \frac{1}{\lambda^2(2^b - 1)}$ and continuing from above we have
\[
\Pr\left[\left|\hG - G(X)\right| \ge \eps_0 (\baG(X) + \rho)\right]
\le 2\exp\left(-\frac{\eps_0^2 m}{(2 + \eps_0)\rho 2^b}(\baG(X) + \rho)\right) \le 2\exp\left(-\frac{\lambda^2 \eps_0^2 m}{4 + 2\eps_0}(\baG(X) + \rho)\right)
\]
where the last inequality follows from $\frac{2^b}{2^b - 1} \le 2$. Letting $\eps = \eps_0 (\baG(X) + \rho)$ we have
\begin{align*}
\Pr\left[\left|\hG - G(X)\right| \ge \eps\right]
&\le 2\exp\left(-\frac{\lambda^2 \eps^2 m}{4\baG(X) + 4\rho + 2\eps}\right).
\end{align*}
Substituting $m = \frac{n}{2}$ and $\baG(X) = 1 - G(X)$, and then rearranging, we have that for all $\eps \ge 0$ and all $\delta \in (0, 1)$ if we want $|\hG - G(X)| \le \eps$ to hold with probability at least $1 - \delta$ then it suffices that
\[
n = \Omega\left(\frac{(1 - G(X) + \rho + \eps)\log \frac1\delta}{\lambda^2 \eps^2}\right)
\]
and since $3\max\{x, y, z\} \ge x + y + z$ it is enough to have
\[
n = \Omega\left(\frac{\max\{1 - G(X), 2^{-b}, \eps\}\log \frac1\delta}{\lambda^4 \eps^2}\right)
\]
where we plugged in $\rho = \frac{1}{\lambda^2 (2^b - 1)}$. Note that if $\alpha \le 1$ then $\frac{1}{\lambda} = O\left(\frac{2^b}{\alpha}\right)$ and therefore it suffices that
\[
n = \Omega\left(\frac{2^{4b}\max\{1 - G(X), 2^{-b}, \eps\}\log \frac1\delta}{\alpha^4 \eps^2}\right)
\]
which proves the first sample complexity bound.

As for the second sample complexity bound, since $C(X) = -\log (1 - G(X))$ we have
\begin{align}
\left|C(X) - \hC\right| &= 
\left|\log (1 - \hG) - \log (1 - G(X))\right| = \left|\log \frac{1 - \hG}{1 - G(X)}\right| \notag\\
&= \left \lvert \log \frac{1 - G(X) + G(X) - \hG}{1 - G(X)} \right\rvert \le \log\left(1 + \frac{\left|\hG - G(X)\right|}{1 - G(X)}\right) \le \frac{\left|\hG - G(X)\right|}{1 - G(X)}. \label{eq:boundonr}
\end{align}
Observe that $G(X) \in [0, 1 - \frac1k]$ for any random variable $X$ with support size $k$. Therefore by Eq.~\eqref{eq:boundonr} we have
\[
|\hC - C(X)| \le \frac{\left|\hG - G(X)\right|}{1 - G(X)} \le k \left|\hG - G(X)\right| \le \eps.
\]
where the last inequality holds if
\[
n = \Omega\left(\frac{2^{4b}k^2\log \frac1\delta}{\alpha^4 \eps^2}\right)
\]
by the first sample complexity bound, because $\max\{1 - G(X), 2^{-b}, \eps\} \le 1$. 
\end{proof}

\end{document}